%% file: main.tex
\documentclass[10pt,twocolumn,letterpaper]{article}

\usepackage{iccv}              


\definecolor{iccvblue}{rgb}{0.21,0.49,0.74}
\usepackage[pagebackref,breaklinks,colorlinks,allcolors=iccvblue]{hyperref}

\def\paperID{983} 
\def\confName{ICCV}
\def\confYear{2025}

\title{MUG: Pseudo Labeling Augmented Audio-Visual Mamba Network for Audio-Visual Video Parsing}

\author{Langyu Wang\quad
Bingke Zhu\thanks{Corresponding author.}\quad
Yingying Chen\quad
Yiyuan Zhang\quad
Ming Tang\quad
Jinqiao Wang \\
$^1$ Foundation Model Research Center, Institute of Automation, \\ 
Chinese Academy of Sciences, China \\
$^2$ School of Artificial Intelligence, University of Chinese Academy of Sciences, China\\
{\tt\small \ wangly54321@163.com \quad
             \{bingke.zhu, yingying.chen, yiyuan.zhang, tangm, jqwang\}@nlpr.ia.ac.cn}}

\begin{document}
\maketitle
\input{sec/0_abstract}    
\input{sec/1_intro}
\input{sec/2_formatting}
\input{sec/3_finalcopy}
{
    \small
    \bibliographystyle{ieeenat_fullname}
    \bibliography{main}
}

\end{document}

%% file: sec/0_abstract.tex
\begin{abstract}
The weakly-supervised audio-visual video parsing (AVVP) aims to predict all modality-specific events and locate their temporal boundaries. Despite significant progress, due to the limitations of the weakly-supervised and the deficiencies of the model architecture, existing methods are lacking in simultaneously improving both the segment-level prediction and the event-level prediction. In this work, we propose an audio-visual \textbf{M}amba network with pseudo labeling a\textbf{UG}mentation (MUG) for emphasising the uniqueness of each segment and excluding the noise interference from the alternate modalities. Specifically, we annotate some of the pseudo-labels based on previous work. Using unimodal pseudo-labels, we perform cross-modal random combinations to generate new data, which can enhance the model’s ability to parse various segment-level event combinations. For feature processing and interaction, we employ an audio-visual mamba network. The AV-Mamba enhances the ability to perceive different segments and excludes additional modal noise while sharing similar modal information. Our extensive experiments demonstrate that MUG improves state-of-the-art results on LLP dataset in all metrics ( \textit{e.g.}, gains of 2.1\% and 1.2\% in terms of visual Segment-level and audio Segment-level metrics). Our code is available at \url{https://github.com/WangLY136/MUG}.
\end{abstract}

%% file: sec/1_intro.tex
\section{Introduction}
\label{sec:intro}
\indent 
\begin{figure} [t!]
	\centering
	\subfloat[\label{1a}]{
		\includegraphics[width=0.5\textwidth]{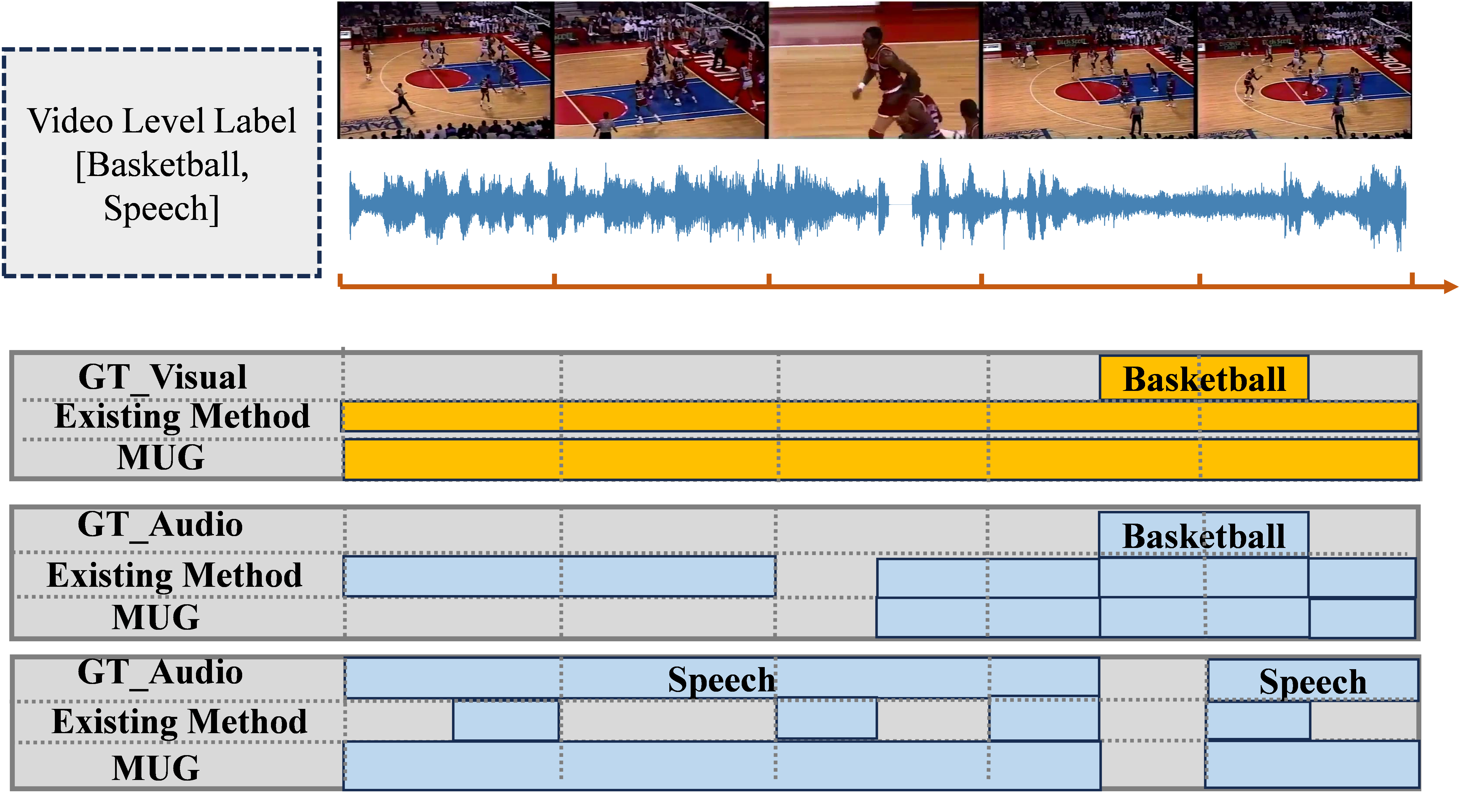}}
	\\
	\subfloat[\label{1b}]{
		\includegraphics[width=0.5\textwidth]{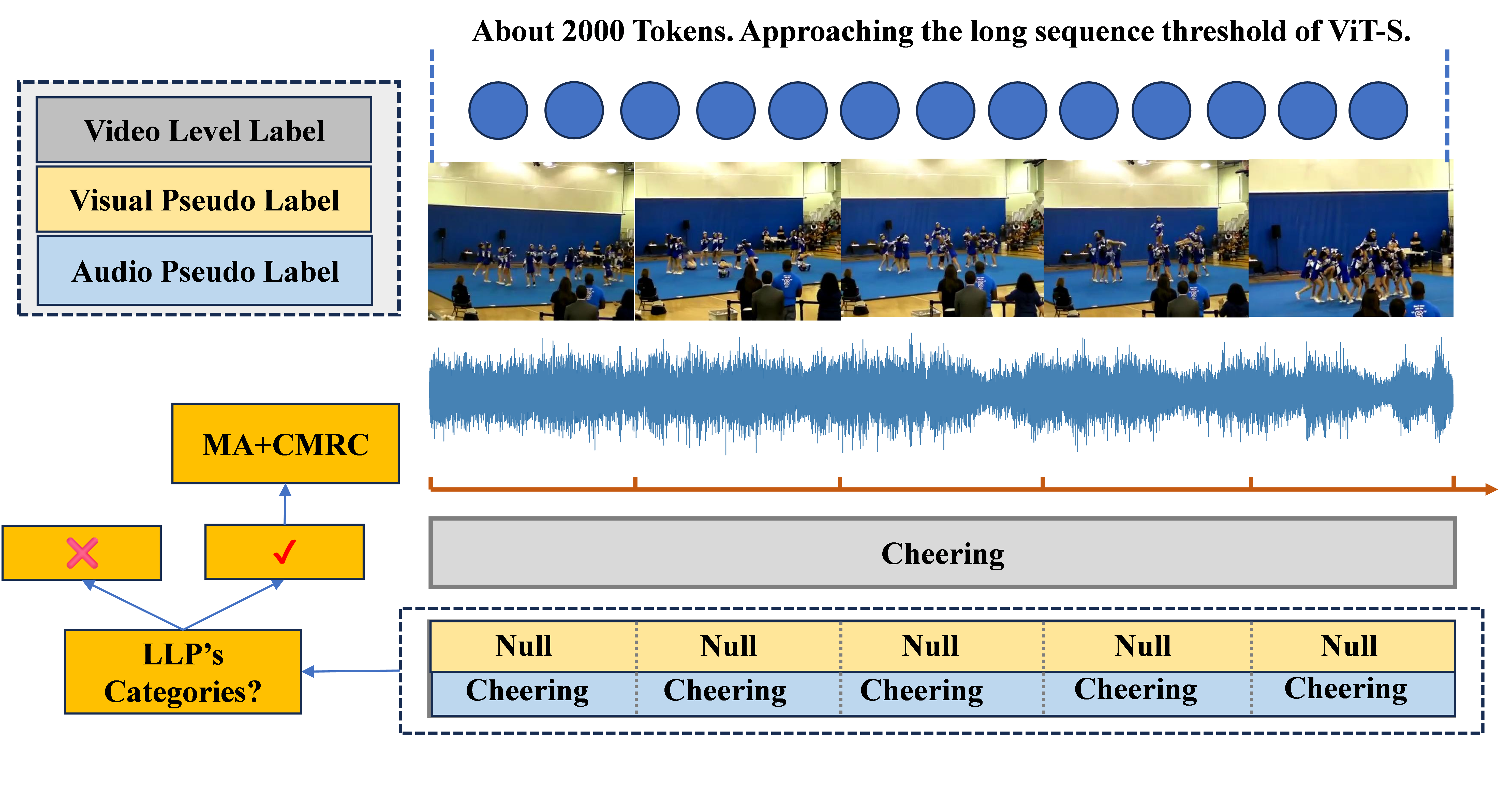} }
	\caption{\footnotesize (a): With only video level label, AVVP requires parsing out visual events, audio events, and their temporal boundaries. (b): Constrained by the weakly-supervision learning and the noise in LLP dataset, previous works fail to learn a large number of event combinations at the segment-level, and a large quantity of empty pseudo-labels occur. Meanwhile, the input tokens are approaching the long sequence threshold of ViT-S \cite{forty-seven}.}
	\label{fig3} 
\vspace{-2.0em}
\end{figure}

Multimodal learning is now a crucial field in machine learning. Many audio-visual tasks such as audio-visual event localization  \cite{one} and audio-visual question answering \cite{two} assume that modalities are aligned and both visual and audio modalities contain learnable cues. However, in the real world, audio-visual events are often unaligned, \textit{e.g.}, seeing a group of people playing basketball but hearing the honking of cars on the road. After observing this prevalent modal mismatch, Tian \textit{et al.} \cite{three} proposed the audio-visual video parsing task. The aim of audio-visual video parsing is to classify video events and localize them according to time and modality. Due to the laborious labeling process, the LLP dataset proposed by Tian \textit{et al.} is only trained using weakly-supervised approach, where only video-level labels are provided for the whole training process (\ref{1a}).

Limited by weakly-supervised learning, the models can only learn the features of different events from video-level labels. Although previous  works \cite{seven, forty-six} have been devoted to extracting pseudo-labels, the quality of the pseudo-labels is restricted due to the presence of noise in the labels of the LLP dataset. Existing frameworks fail to learn a large number of event combinations at the segment-level, which significantly affects the model's prediction of segment-level events. Meanwhile, the events occurring in each modality are independent of each other and may even be completely unrelated. An event in one modality may either complement the prediction of another modality or introduce noise. Some works \cite{thirteen, ten, eleven, fifty-one} attempt to develop more robust audio-visual encoders for embedding more effective audiovisual features, but they are insufficient in retaining the original modality features while sharing the features. Finally, the AVVP task requires the simultaneous prediction of events at both the Segment-level (single frame) and the Event-level (multiple frames). Therefore, it is necessary not only to conduct inference on single-frame images/audio but also to capture the strong causal relationships among multiple frames of images/audio. The commonly used Transformer (primarily due to its attention mechanism) models in the past have certain limitations when dealing with long sequences (a large number of tokens). The token length of the visual input sequence in AVVP is close to 2000, which is approaching the threshold of ViT-S \cite{forty-seven}. Recently, some Mamba-based backbones \cite{eight, nine} have demonstrated great potential in long-sequence modeling and can utilize the causal order to model sequences. However, for single-frame image, there is no sequential dependency. Instead, the ability to model the overall space is required \cite{forty-seven, forty-eight}. Therefore, Mamba has deficiencies in the recognition of single-frame image.

In order to effectively enhance the model's perception ability of segment-level features, we propose a brand-new data augmentation strategy applicable to the AVVP task. We extract all the pseudo-labels from previous works \cite{seven} and manually annotate the obviously incorrect pseudo-labels (empty labels) among them. Subsequently, we extract the visual track of one video and the audio track of another video, and randomly combine them into a new video. The label of new video is the intersection of the original visual pseudo-label and the auditory pseudo-label. It is worth noting that the pseudo-labels that cannot be annotated and their corresponding videos will be discarded during the random combination. At the same time, we introduce the text modality into the AVVP task. We extract the semantic information of the pseudo-labels and adaptively fuse it with the visual/audio features to eliminate additional modal noise.

Inspired by the Mamba-Transformer architecture in previous works \cite{forty-eight, forty-nine, fifty}, based on the HAN \cite{three} model, we propose a brand-new baseline that can simultaneously improve the model's performance at both the segment-level and the event-level. Specifically, we first utilize a Mamba-based attention to capture the key information in the sequence. Subsequently, we propose a cross-modal adaptive mamba fusion structure which can captures the cross-modal similar information while retaining the intermodal information through a shared matrix. In order to prevent the causal model from forgetting the early token information, we add an additional dynamic branch to alleviate this problem. To enhance the cross-modal similar information captured in the previous steps, we introduce a Mamba feature enhancement module. We incorporate the HAN model at the end of the network, and its simple Transformer (attention architecture) further strengthens the long-range spatial dependencies. 

Extensive experimental results on LLP data demonstrate that MUG outperforms existing state-of-the-art models on several metrics. Compared with pure Transformer or CNN architectures, our method achieves more advanced results. Our contributions are summarized as follows:

\begin{itemize}
    \item We propose a data augmentation approach to effectively improve the model's prediction ability for segment-level, and it can be applied to multiple downstream models;
    \item We investigate a Mamba-Transformer network, which simultaneously improves the detection accuracy of the model at both the segment-level and the event-level;
    \item Text features are introduced to exclude the noise of another modality and constrain the prediction of unimodal.
\end{itemize}

\section{Related Works}
\textbf{Audio-Visual Video Parsing (AVVP).} The aim of audio-visual video parsing is to identify visual and audio events in a video and locate their timestamps under weakly supervised conditions. Tian \textit{et al.} \cite{three} first introduced the AVVP task and proposed a framework based on hybrid attention networks and multimodal multi-instance learning. Based on this, numerous studies have focused on network architecture construction and the application of attention mechanisms. Yu \textit{et al.} \cite{ten} proposed a multimodal pyramidal attention network for capturing and integrating multilevel features. Mo \textit{et al}. \cite{eleven} proposed a multimodal grouping network to learn dense and differentiated audio-visual encodings. Wu \textit{et al.} \cite{twelve} designed an algorithm to obtain modality-related labels by exchanging audio and visual tracks. Duan \textit{et al.} \cite{thirteen} used a bi-directionally guided multi-dimensional attentional mechanism to improve performance on a variety of downstream tasks. Gao \textit{et al.} \cite{fourteen} proposed a joint modal mutual learning process that adaptively and dynamically calibrated the evidence for a variety of audible, visible, and audible-visible events. In addition, there are some works that use label denoising strategies or generates pseudo-labels for finer-grained supervised learning. Cheng \textit{et al.} \cite{fifteen} dynamically identified and removed modality-specific noisy labels in a two-phase approach. Lai \textit{et al.} \cite{seven} used frozen CLIP \cite{four} and CLAP \cite{five} to extract features and generated pseudo-labels to aid prediction. Fan \textit{et al.} \cite{forty-six} proposed to perform dynamic re-weighting method to adjust the pseudo-labels. Zhou \textit{et al.} \cite{seventeen} built on pseudo-labels to propose a novel decoding strategy to solve the problem of parsing potentially overlapping events. However, the above methods do not address the limitations inherent in the LLP dataset, which leads to insufficient accuracy of pseudo-labels and insufficient quality of the dataset. Previous works also faced the problem of introducing noise from another modality.

\noindent \textbf{Data Augmentation.} Data augmentation techniques are crucial in model training, enhancing generalization and robustness, especially when data is limited. There have been many very mature and widely used data augmentation methods for images, for example \cite{eighteen, nineteen, twenty, twenty-one, twenty-two, twenty-three}. However, the application of data augmentation in the video domain remains relatively sparse. Kim \textit{et al.} \cite{twenty-four} extended the data augmentation strategy for images to the temporal dimension of videos as a way to learn temporal features in videos. Yun \textit{et al.} \cite{twenty-five} extended CutMix in the field of image recognition to the video domain by proposing VideoMix. Zhang \textit{et al.} \cite{twenty-six} observed the effect of hue changes on video understanding and proposed a data enhancement method called motion related enhancement. Building on this foundation, we propose a pseudo-label-based cross-modal random combination method for AVVP task, which can effectively improve the generalization and robustness of the model.
\begin{figure}[h]
\vspace{-1.0em}
    \centering
    \includegraphics[height=3cm, width=\linewidth]{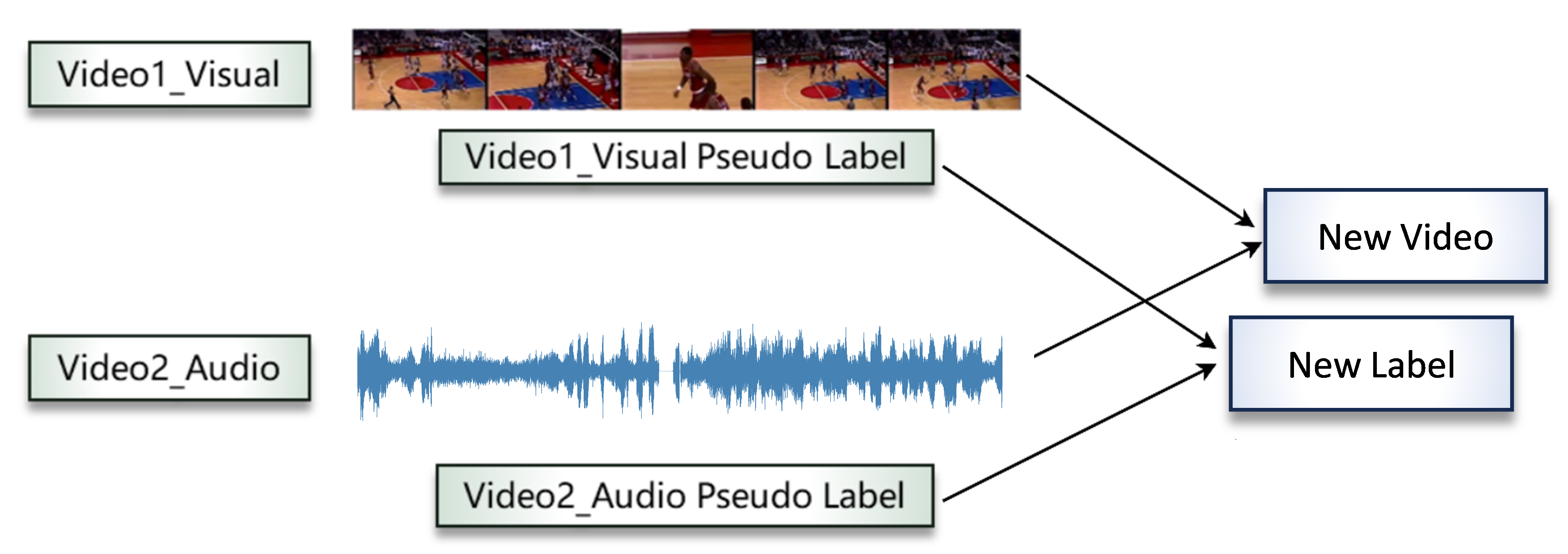}
    \caption{\footnotesize The process of the CMRC.}
    \label{figure2}
\vspace{-1.0em}
\end{figure}

\noindent \textbf{State Space Models (SSM).} The state space model \cite{twenty-seven} has become one of the most important backbones in deep learning. It originates from classical control theory and provides linear scalability of sequence length for modeling remote dependency. To enhance the practical feasibility, Gu \textit{et al.} \cite{twenty-eight} proposed the S4 module. Subsequently, Smith \textit{et al.} \cite{twenty-nine} proposed a SSM model supporting multiple inputs and multiple outputs, and Hasani \textit{et al.} \cite{thirty} proposed a SSM model for liquid structures. Based on this, Gu \textit{et al.} \cite{eight} proposed the mamba architecture. It merges the previous SSM structure with the MLP module in Transformer into a single module, thus obtaining an architectural design with a selective state space. Inspired by ViT \cite{thirty-one} and Swin Transformer \cite{thirty-two}, Zhu \textit{et al.} \cite{thirty-three} and Liu \textit{et al.} \cite{thirty-four} proposed Vision Mamba and VMamba, respectively, which showed impressive potential on numerous vision tasks. Meanwhile, Mamba structures have begun to emerge on multimodal tasks. xie \textit{et al.} \cite{thirty-five} proposed a mamba multimodal fusion structure for medical images. Li \textit{et al.} \cite{thirty-six} utilised a coupled state space model to enhance information fusion between different modalities, which greatly improves the inference speed. All these works lay the foundation for subsequent multimodal mamba research.

\begin{figure*}
    \centering
    \includegraphics[height=10cm, width=\linewidth]{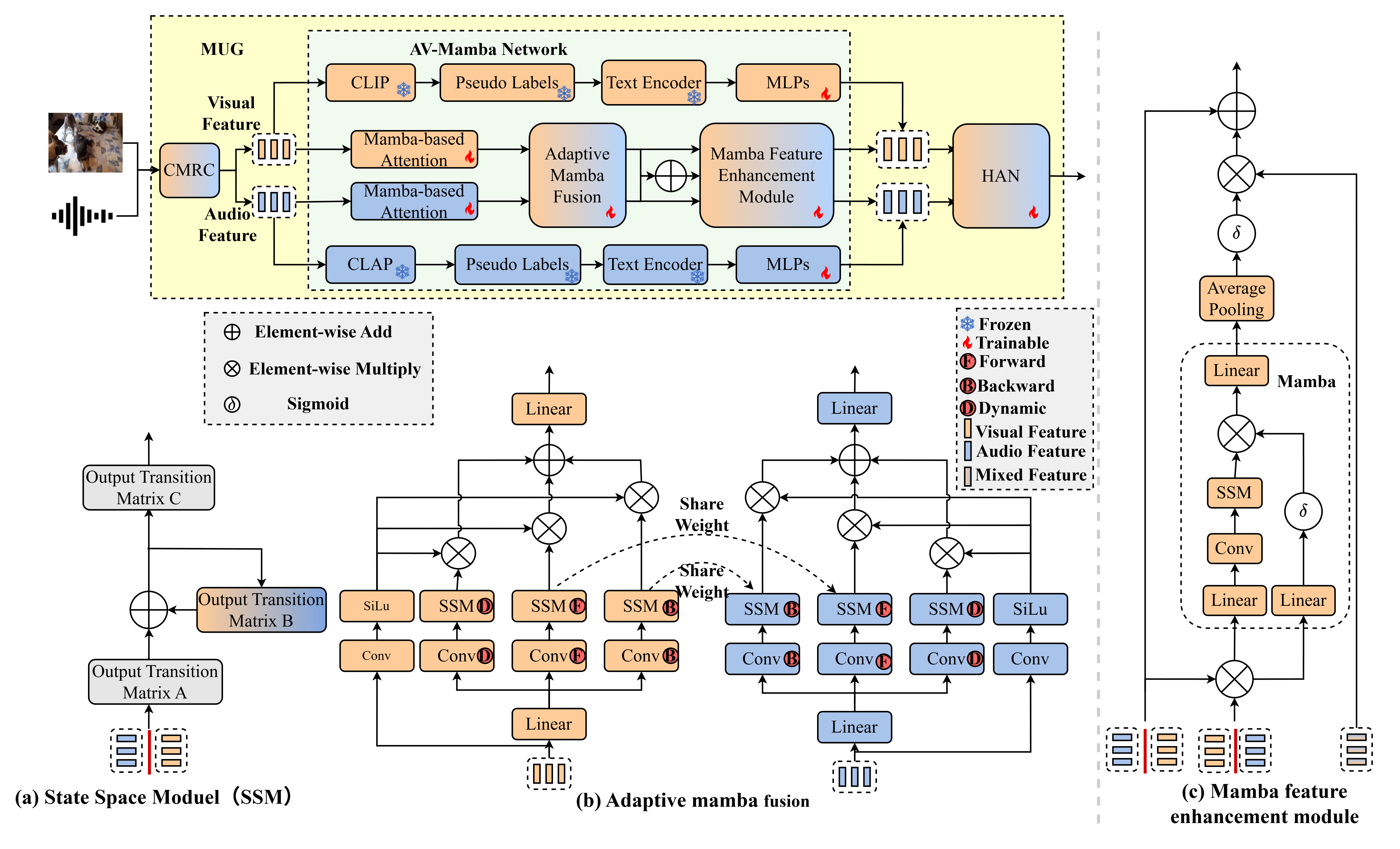}
    \vspace{-2.0em}
    \caption{\footnotesize The framework of \textbf{MUG}. MUG consists of two parts: the data augmentation CMRC and the AV-Mamba Network. Pseudo-labels are extracted by VALOR \cite{seven}, which can be used to provide fine-grained supervision and extract semantic features by CLIP/CLAP .}
    \label{figure3} 
    \vspace{-2.0em}
\end{figure*}

\section{Our Approach}
We make improvements from both the data and model perspectives. On the data side, we propose a data augmentation method specifically for the AVVP task. We first annotate some obviously erroneous pseudo-labels. Building on this, we randomly combine the visual and audio modalities of any two videos to generate new data. This method effectively enhance the quality of the dataset, allowing the model to learn the features of each segment more thoroughly. On the model side, we propose an audio-visual mamba network. We capture both temporal and local information simultaneously using Mamba-based attention and share some parameters of the SSM. This approach not only enhances the model's perception of each segment but also shares cross-modal similar information while preserving unimodal information. For effective projection and better model optimization, we incorporate the
segment-wise pseudo labels generated in recent work \cite{twenty-one} to provide fine-grained supervision. We also introduce the text modality to exclude irrelevant modality noise. The MUG framework is shown in \ref{figure3}. We describe the problem formulation in section 3.1. Then we illustrate the data augmentation and mamba framework in section 3.2 and section 3.3. 

\subsection{Problem Definition}
The AVVP task aims to identify the event of every segment into audio event, visual event and audio-visual event, together with their classes. For the benchmark dataset of Look, Listen, and Parse (LLP), a T-second video is split into T non-overlapping segments, expressed as S = $\lbrace A_t,V_t\rbrace ^T_{t=1}$, where A and V represent audio and visual segment in time t respectively. In each segment, $y^a_t \in \mathbb R^C$, $y^v_t \in \mathbb R^C$, $y^{av}_t \in \mathbb R^C$ represent to the audio event labels, visual event labels and audio-visual event labels, $C$ is the number of event types. However, we only have weak labels in training split, but have detailed event labels with modalities and temporal boundaries for evaluation.

\subsection{Data augmentation in AVVP}
\textbf{Manual annotation (MA).} Lai \textit{et al.}\cite{seven} used frozen CLIP and CLAP to compute the visual segment pseudo-labels $\hat{y}^m_v$ and audio segment pseudo-labels $\hat{y}^m_a$, respectively. In process of generating pseudo-labels, Lai \textit{et al.} use the pre-trained CLIP and CLAP to compute the cosine similarity of the features and the 25 categories of labels, and pass a threshold to obtain one-hot encoded pseudo-labels.  However, the LLP dataset itself is not annotated sufficient accuracy and the 25 categories cannot encompass all events. The limitation of the LLP  dataset is one of the reasons that lead to the insufficient accuracy of pseudo-labels. We convert the generated one-hot pseudo-labels to CSV file, and we can find that there are many null labels (\ie, no events occurred) in the visual modality, which is abnormal since a video may lack sound but rarely lacks image. For such pseudo-labels, we compare them with the LLP dataset and manually annotate them. Pseudo-labels that belong to the 25 categories are accurately annotated, while those outside this range are left unannotated and excluded during the cross modality random combination process.

\noindent \textbf{Cross modality random combination (CMRC).} The inability to use data enhancement in previous works is due to the absence of unimodal label information. With the availability of high-quality pseudo-labels, data enhancement for the AVVP task became feasible. For the LLP dataset, we first count the distribution of all video-level labels that appear more than 50 times in the entire training set. Based on this distribution, we selectively combine the visual features of one video with the audio features of another video to generate new video data. The label of the new video is the union of the visual modality pseudo-labels and the audio modality pseudo-labels. \ref{figure2} represents the process of the CMRC. The pseudo-labels that cannot be annotated and their corresponding videos (\ie, pseudo-labels and videos that do not fall within the 25 categories) are eliminated when generating new data. In order to avoid overfitting and introduce excessive additional data noise, we generate five batches of data according to the actual distribution, with quantities of 1585, 3242, 4610, 6080 and 12096 respectively. We separately test data augmentation for different batches to find the critical point of this method. When the amount of generated data is small, there is still room for model optimization. When the generated data is abundant, the noise in the dataset will increase, and meanwhile, the risk of overfitting rises. CMRC is reasonable because in the real world, visual and audio modalities are not always correlated, because sound signals can originate from various directions. 

\subsection{AV-Mamba Network}

\noindent \textbf{Mamba-based attention (MBA).} Audio and visual features are extracted using pre-trained VGGish and ResNet-152, denoted as $\lbrace f^a_t \rbrace ^T_{t=1}$, $\lbrace f^v_t \rbrace ^T_{t=1}$. $F^a$ and $F^v$ stand for the feature set in the same video, which are defined as $F^a =\lbrace f^a_1, ..., f^a_T \rbrace \in \mathbb R^{T\times d}$ and $F^v =\lbrace f^v_1, ..., f^v_T \rbrace \in \mathbb R^{T\times d}$, $d$ is the feature dimension. To aggregate features from different segments and enhance the feature expressiveness of each segment, we propose a segment-based  attention module implemented via Mamba. This module is similar to the convolutional block attention module \cite{thirty-seven} but it is implemented by Mamba, thereby better handling the causal relationship.

For each segment, two feature vectors are generated by global maximum pooling and global average pooling. These feature vectors are fed into a shared fully connected layer, resulting in the final attention weight vector $W^m_t$:
\begin{equation}
\resizebox{0.9\hsize}{!}{$
    W^m_t(f^m_t) = \delta (Mamba(AvgPool(f^m_t))+Mamba(MaxPool(f^m_t ))),
$}
\end{equation}
where $f^m_t$ denotes audio or visual features within a video, whose dimensions are $(B, T, D)$, representing batchsize, segments, and 128/2048 (audio/visual), respectively. $\delta$ indicates sigmoid function, $m \in \{a,v\}$. $Mamba$ denotes Mamba block.

Similar to aforementioned method, two feature vectors are produced through max pooling and average pooling along the temporal (segment) dimension. The features derived from max pooling and average pooling are then concatenated along the temporal dimension to obtain a feature vector $S^m_t$, $m \in \{a,v\}$:
\begin{equation}
\label{eq3}
\resizebox{0.9\hsize}{!}{$
    S^m_t(f^m_t) = \delta (Mamba(AvgPool(f^m_t));(MaxPool(f^m_t ))).  
$}
\end{equation}
 The outputs $W^m_t$ and $S^m_t$ are element-wise multiplied to obtain the final attention-enhanced feature $\hat{f}^m_t$, $m \in \{a,v\}$:
\begin{equation}
    \hat{f}^m_t =S^m_t(W^m_t(f^m_t) \otimes f^m_t)\cdot (W^m_t(f^m_t) \cdot f^a_t).
\end{equation}

\noindent \textbf{Adaptive mamba fusion (AMF).} In previous work \cite{thirty-eight, thirty-nine, forty}, there have been explorations in multimodal Mamba fusion with surprising results. Building upon these findings, we propose an adaptive mamba fusion network for selective interaction of visual and audio features. Specifically, after Mamba-based attention, visual features and audio features are fed into AMF. Each of the two input features in the AMF is processed through four distinct branches, which capture unique features. $f^{Forward}_m$ is forward process, performing forward scans. $f^{Backward}_m$ is backward process, performing backward scans. The forward and backward scanning can be expressed as:
\begin{equation}
    f^{Forward}_m = SSM([\hat{f}^m_1,\hat{f}^m_2,...,\hat{f}^m_N], Conv1d),
 \end{equation} 
\begin{equation} 
    f^{Backward}_m = SSM([\hat{f}^m_N,\hat{f}^m_{N-1},...,\hat{f}^m_1], Conv1d),
\end{equation}
where $Conv1d$ represents 1D convolution, $SSM$ represents the state space model, $m \in \{a,v\}$. However, only performing forward and backward scans can potentially ignore the first and last layer tokens. The Mamba has some risk of forgetting the initial token entered, so relying only on forward and backward is not adequate way to model all layers of features. For the input features, the importance of each segment is different and the events that occur are different. We have incorporated the dynamic scanning scheme  from recent work \cite{forty}, allowing the scanning process to start and end at any layer. This approach enhances the model's comprehension of different segments and thus improves the ability to parse segment-level events. 

Furthermore, in the forward/backward SSM, two modalities share a portion of parameters due to the consistency in their scanning directions. As shown in \ref{figure3}, forward SSM and backward SSM include 6 parameters: input transition matrices A state transition matrices B,and output transition matrices C. The state transition matrices B has the most significant impact on the system, as they govern the evolution of the current hidden state . Consequently, We share parameters of state transition matrices B between two modalities, while keeping input transition matrices a and output transition matrices C independent \cite{thirty-eight}. This strategy not only reduces the number of parameters and the potential risk of overfitting, but also preserves unimodal independence while capturing cross-modal similarity information.  Subsequently, each feature of both modalities is subjected to a gating strategy in order to fuse the scanning results and obtain the feature $f^{AMF}_m$ by element-by-element summation, $m \in \{a,v\}$. Finally, as shown in \ref{figure3}, the features undergo a simple add to obtain mixed feature $f^{AMF}_{mix}$. $f^{AMF}_m$ and $f^{AMF}_{mix}$ are fed as three input to the Mamba feature enhancement module. We employ multiple learnable parameters to control the degree of interaction between different modalities, preventing excessive interference between modalities. By modeling sequences in causal order, Mamba strengthens the connections between adjacent frames and enhance the perception ability of multi-frame events.

\noindent \textbf{Mamba feature enhancement module(MFE).} The features that have undergone AMF will be fed into the mamba feature enhancement module, as illustrated in \ref{figure3}. This module accepts three types of input: visual features, audio features, and mixed features. Initially, the feature maps from the two modalities undergo average pooling to reduce their dimensionality. Next, the features from the corresponding time steps of the two modalities are subjected to an element-wise multiplication operation. These features are then transformed into channel enhancement vectors through the activation of a Sigmoid function, which performs channel-wise enhancement on the original feature maps at each time step. By leveraging the element-wise product of features from different modalities, the mapping of these similar features is strengthened. This process effectively amplifies the shared features and fine details across the two modalities while suppressing the dissimilar features, thereby mitigating the interference caused by discrepancies between modal features. Finally, we obtain the enhanced features $f^{MFE}_m$, $m \in \{a,v\}$.

\noindent \textbf{Pseudo-label semantic interaction module (PLSIM).} The information in text can be used as a cue to effectively improve the performance of the models \cite{sixteen, seventeen, forty-one, fifty-one}.
The above work has proven that text can serve as a priori information to culling out the other modal noise and constrain uni-modal event prediction. Contrary to the pseudo-labels employed in \cite{seven} , we eliminate logical operations with video-level labels when generating pseudo-labels for testing purposes. New pseudo-labels can be expressed as $\hat{y}^m_t$.

Since both pseudo-labels $\hat{y}^m_t$ and real labels $y$ are one-hot coded, the event categories corresponding to pseudo-labels $f^a_{event}$ and $f^v_{event}$ can be easily extracted. Next, we convert the event categories in the pseudo-labels into concepts that can be understood by CLIP/CLAP. The title for each event is formulated by prepending the prefix ‘A photo of’ or ‘this is a sound of’ to the natural language description of the event. These captions are processed by a frozen CLIP/CLAP text encoder to obtain pseudo-label semantic features $F^a_{CLAP}$ and $F^v_{CLIP}$ for linguistic consistency: 
\begin{equation}
\resizebox{0.9\hsize}{!}{$
    F^a_{CLAP} = CLAP(f^a_{event}), F^v_{CLIP} = CLIP(f^v_{event}).
$}
\end{equation}
 
  In addition, we use multiple MLPs $\varDelta^n_m$ to map the semantic information of the text, which can be written as:
\begin{equation}
    \gamma_{a1} = \varDelta^1_m(F^a_{CLAP}),\gamma_{a2} = \varDelta^2_m(F^a_{CLAP}),
\end{equation}
\begin{equation}
    \rho_{v1} = \varDelta^3_m(F^v_{CLIP}),  \rho_{v2} = \varDelta^4_m(F^v_{CLIP}),
\end{equation}
where $\varDelta^1_m$, $\varDelta^2_m$, $\varDelta^3_m$, $\varDelta^4_m$ are different MLPs operations to generate the semantic parameters respectively. We use the extracted audio/visual features to fuse with the semantic information, which can be represented as:
\begin{equation}
    F_{audio} = \ f^{MFE}_a \odot \gamma_{a1} + \gamma_{a2} + \ f^{MFE}_a,
\end{equation}
\begin{equation}
    F_{visual} = \ f^{MFE}_v \odot \rho_{v1} + \rho_{v2}+ \ f^{MFE}_v,
\end{equation}
where $\odot$ denotes Hadamard product. $\gamma_{a1}$ and $\rho_{v1}$ denote scale scaling, $\gamma_{a2}$ and $\rho_{v2}$ denote bias control. $F_a$ and $F_v$ are audio and visual features fused with semantic features. It is worth noting that when generating new data using CMRC, PLSIM will also to carry out expansion to achieve the fusion of the corresponding features. In the experiments, we find that when using batches augmented with 12,000 samples, the improvement of the PLSIM module decreased significantly and even produced negative effects. This may be due to the fact that excessive data augmentation introduces excessive label noise, thereby affecting the performance of the PLSIM module.

\section{Experiments}
\subsection{Experimental setup}

\textbf{LLP Dataset.} The LLP dataset \cite{three} is used to evaluate our method. This dataset has 11849 videos with 25 categories taken from YouTube and consist of a wide variety of scene content including daily activities, music performances, vehicle sounds etc. The dataset has 10000 videos with weak labels as the training set, 1200 videos and 649 videos as the testing set and the validation set with fully annotated labels.

\noindent \textbf{Evaluation Metrics.} Following previous works, we use F1-
scores on audio, visual and audio-visual events as evaluation metrics. These are computed both at segment and event level. We also include the aggregate metrics “Type@AV” and “Event@AV”, again compute at the segment and event level. See \cite{three} for a full explanation of metrics.

\noindent \textbf{Implementation Details.} We conduct the training and evaluation processes on a NVIDIA RTX A6000 GPU with 48GB memory. Following the data preprocessing in previous works, we decode a 10-second video at 8 fps into 10 segments. Audio input tokens are extracted through
pre-trained VGGish \cite{forty-three}, and visual tokens are obtained through the pre-trained models ResNet-152 \cite{forty-two} and R(2+1)D. Our model is trained using Adamw with batchsize of 64 and a learning rate of $3e^{-4}$ for 20 epochs.
\begin{table*}[]
\centering
\label{tab:my-table}
\resizebox{\textwidth}{!}{
\begin{tabular}{c|c|ccccc|ccccc}
\hline
                                      &                                   & \multicolumn{5}{c|}{\textbf{Segment-level}}                                                                                                         & \multicolumn{5}{c}{\textbf{Event-level}}                                                                                                            \\ \cline{3-12} 
\multirow{-2}{*}{\textbf{Method}}     & \multirow{-2}{*}{\textbf{Venue}}  & \textbf{A}                  & \textbf{V}                  & \textbf{AV}                 & \textbf{Type@AV}            & \textbf{Event@AV}           & \textbf{A}                  & \textbf{V}                  & \textbf{AV}                 & \textbf{Type@AV}            & \textbf{Event@AV}           \\ \hline
HAN\cite{three}                            & ECCV’20                           & 60.1                        & 52.9                        & 48.9                        & 54.0                        & 55.4                        & 51.3                        & 48.9                        & 43.0                        & 47.7                        & 48.0                        \\
MM-Pyr\cite{ten}                          & MM’22                             & 60.9                        & 54.4                        & 50.0                        & 55.1                        & 57.6                        & 52.7                        & 51.8                        & 44.4                        & 49.9                        & 50.0                        \\
MGN\cite{eleven}                          & NeurIPS’22                        & 60.8                        & 55.4                        & 50.4                        & 55.5                        & 57.2                        & 51.1                        & 52.4                        & 44.4                        & 49.3                        & 49.1                        \\
JoMoLD\cite{fifteen}                        & ECCV’22                           & 61.3                        & 63.8                        & 57.2                        & 60.8                        & 59.9                        & 53.9                        & 59.9                        & 49.6                        & 54.5                        & 52.5                        \\
CMPAE\cite{fourteen}                       & CVPR’23                           & 64.2                        & 66.4                        & 59.2                        & 63.3                        & 62.8                        & 56.6                        & 63.7                        & 51.8                        & 57.4                        & 55.7                        \\
DGSCT\cite{thirteen}                        & NeurIPS’23                        & 59.0                        & 59.4                        & 52.8                        & 57.1                        & 57.0                        & 49.2                        & 56.1                        & 46.1                        & 50.5                        & 49.1                        \\
 VALOR\cite{seven}    & NeurIPS’23    &  61.8         &  65.9     &  58.4         &  62.0         &  61.5         &  55.4         &  62.6         &  52.2         &  56.7         &  54.2 \\
CM-PIE\cite{forty-four}                        & ICASSP’24          & 61.7                       & 55.2                        & 50.1                        & 55.7                       & 56.8                        & 53.7                        & 51.3                        & 43.6                       & 49.5                        & 51.3                        \\
LEAP\cite{seventeen}                        & ECCV’24                           & 62.7                        & 65.6                        & 59.3                        & 62.5                        & 61.8                        & 56.4                        & 63.1                        & 54.1                        & 57.8                        & 55.0                        \\
CoLeaF\cite{forty-five}                      & ECCV’24                           & 64.2                        & 64.4                        & 59.3                        & 62.6                        & 62.5                        & 57.6                        & 63.2                        & 54.2                        & 57.9                        & 55.6                        \\
 \hline
\textbf{MUG}                         & \textbf{-}                        & \textbf{65.4}               & \textbf{66.5}               & \textbf{59.9}               & \textbf{63.9}               & \textbf{64.7}               & \textbf{59.5}               & \textbf{63.9}              & \textbf{55.3}               & \textbf{59.6}               & \textbf{57.7}               \\ 
\hline
\end{tabular}
}
\vspace{-0.5em}
\caption{\footnotesize Comparison with the state-of-the-art methods on the LLP dataset in terms of F-scores. }

\label{table1}
\vspace{-1.5em}
\end{table*}

\subsection{Comparisons with Prior Work}

\noindent \textbf{Quantitative.} We compare our method with several popular baselines, such as HAN, MM-Pyramid, VALOR, DG-SCT in the same dataset (\ref{table1}). From the experimental results, we can see that MUG has improved in all metrics. Compared with the previous SOTA model CoLeaF, it achieves improvements in uni-modal performance, \eg, 2.1\% at the Visual Segment-level (66.6\% vs. 64.4\%), 1.2\% at the Audio Segment-level (65.4\% vs. 64.2\%). Meanwhile, the multi-modal performance is also improved, \eg, 0.6\% at the AV Segment-level (59.9\% vs. 59.3\%) and 1.1\% at the AV event level (55.3\% vs. 54.2\%). Meanwhile, in terms of the Event@AV metric, MUG has an improvement of 2.2\% at the Segment-level (64.7\% vs. 62.5\%) and 2.1\% the Event-level (57.7\% vs. 55.6\%).  Compared with our baseline method VALOR, the proposed method can significantly improve the performances on all metrics. 

\begin{figure}
    \centering
    \includegraphics[height=12cm, width=\linewidth]{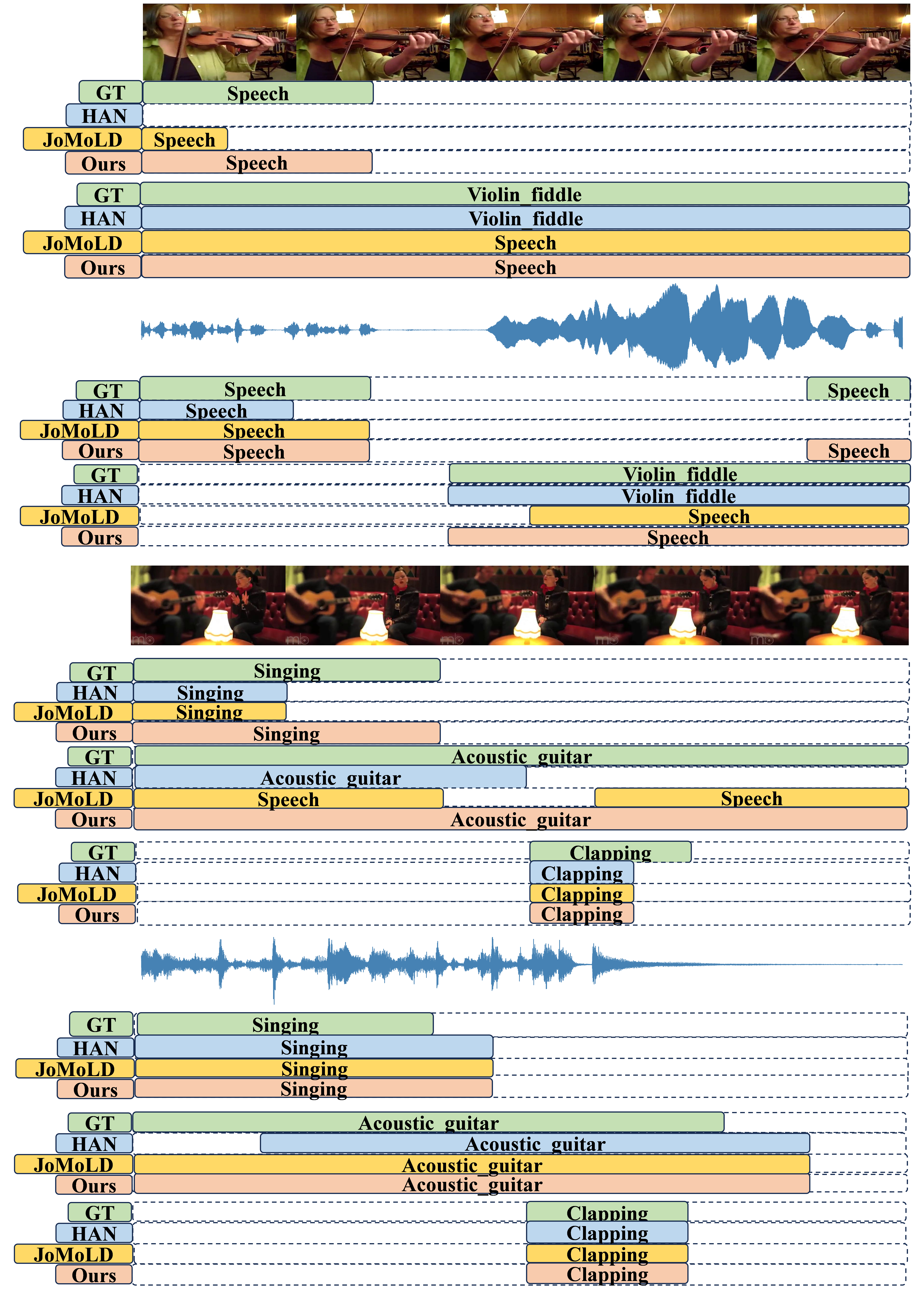}
    \vspace{-1.5em}
    \caption{\footnotesize Examples of Qualitative results.}
    \label{figure4}
\vspace{-2.0em}
\end{figure}

\noindent \textbf{Qualitative.} As shown in \ref{figure4}, we qualitatively compare our method with some previous works. Here we use MUG, HAN, JoMoLD \cite{fifteen} for comparison. In \ref{figure4} above, the first video contains both Speech and Violin and occurs in both modalities. In audio modality, our method localizes the time of Speech that occurs in the last second, while the other two methods fail to do so. In visual modality, although all three methods are accurate in detecting the event Violin, only our method accurately localizes the event Speech. \ref{figure4} (below) presents another video that contains three events, Singing, Guitar and Clapping. In audio modality, our method accurately locates the Singing and Clapping events, with an error of only one second on Guitar. In visual modality, our method accurately locates Singing and Guitar, with only one second error in Clapping. Overall, our method achieves superior parsing results. This indicates that MUG can perceive the events of each segment more accurately and exclude irrelevant interference.

\subsection{Ablation experiments}

\noindent \textbf{Cross modality random combination (CMRC).} Previous work was constrained by weakly supervised learning, which posed challenges for data augmentation. After pseudo-label extraction and manual annotation, we are able to leverage the visual features of one video to randomly combine with the audio features of another video to generate completely new datasets and pseudo-labels. Note that in the random combination process, we excluded all null labels and combined them according to the distribution of the dataset. The ablation experiment of CMRC is shown in \ref{table2}. To facilitate the presentation of the results, we generate five batches of datasets, the results of which are shown in \ref{table3}. The results indicate that training the model with the combined data can effectively enhance accuracy (batch4). There is a significant enhancement in all metrics of visual modality, which is attributed to the fact that obviously wrong pseudo-labels (labels are null) are eliminated during random combination. Due to the limitations of the dataset itself, generating too much data may lead to overfitting, which prevents the model metrics from increasing or even decreasing. Fitting random combination provides a high-quality dataset, enabling the model to thoroughly learn the features of each segment and deepen its ability to perceive each segment. We apply this data augmentation method to a variety of different baselines to demonstrate the effectiveness of CMRC, as shown in \ref{table4}.
\begin{table*}[h]
\center
\begin{tabular}{c|ccccc|ccccc}
\hline
\multirow{2}{*}{Method} & \multicolumn{5}{c|}{Segment-level}         & \multicolumn{5}{c}{Event-level}                                                                                     \\ \cline{2-11} 
                        & A     & V     & AV    & Type@AV & Event@AV & \multicolumn{1}{c}{A} & \multicolumn{1}{c}{V} & \multicolumn{1}{c}{AV} & \multicolumn{1}{c}{Type@AV} & Event@AV \\ \hline
MUG                     & \textbf{65.4}  & \textbf{66.5}  & \textbf{59.9}    & \textbf{63.9}    & \textbf{64.7}     & \textbf{59.5}                     & \textbf{63.9}                   & \textbf{55.3}                    & \textbf{59.6}                         & \textbf{57.7}     \\
wo/CMRC                  & 62.7  & 65.2  & 58.6  & 62.2    & 62.2     & 56.5                   & 61.8                   & 53.8                    & 57.4                         & 54.6     \\
wo/MBA                  & 64.5  & 66.1  & 59.4  & 63.3    & 63.9     & 58.3                   & 63.0                   & 53.9                    & 58.4                         & 56.5     \\
wo/AMF                  & 64.1  & 66.3  & 59.7  & 63.4    & 63.4     & 58.1                   & 63.0                   & 54.8                    & 58.6                         & 56.1     \\
wo/MFE                  & 63.8  & 64.6  & 58.8  & 62.4    & 62.8     & 57.3                   & 61.6                   & 53.5                    & 57.5                         & 55.1   \\
wo/PLSIM                & 64.8 & 66.5 & 59.6 & 63.7    & 64.0    & 58.3                  & 63.3                  & 53.4                   & 58.4                        & 56.3    \\ \hline
\end{tabular}
\vspace{-0.5em}
\caption{\footnotesize Ablation experiments of MUG. wo/ denotes without.}
\label{table2}
\vspace{-1.5em}
\end{table*}

\begin{table}[]
\begin{tabular}{c|ccccccc}
\cline{1-6}
\multirow{2}{*}{Method} & \multicolumn{5}{c}{Segment-level}                 &  &  \\ \cline{2-6}
 & A    & V    & AV   & Type@AV & Event@AV &  &  \\ \cline{1-6}
Batch1 & 63.8 & 65.4 & 58.9 & 62.7    & 62.9     &  &  \\
Batch2 & 64.4 & 65.8 & 59.5 & 63.2    & 63.3     &  &  \\
Batch3 & 64.4 & 66.1 & 59.2 & 63.2    & 63.8     &  &  \\
Batch4 & 65.4 & \textbf{66.5} & \textbf{59.9} & \textbf{63.9}    & 64.7     &  &  \\
Batch5 & \textbf{66.3} & 65.8 & 59.6 & 63.9    & \textbf{64.9 }    &  &  \\ \cline{1-6}
\multirow{2}{*}{Method} & \multicolumn{5}{c}{Segment-level}                 &  &  \\ \cline{2-6}                  
 & A    & V    & AV   & Type@AV & Event@AV &  &  \\ \cline{1-6}
Batch1 & 57.3 & 62.5 & 53.6 & 57.8    & 55.5     &  &  \\
Batch2 & 58.3 & 62.4 & 53.9 & 58.2    & 56.0     &  &  \\
Batch3 & 58.2 & 62.8 & 53.7 & 58.2    & 56.3     &  &  \\
Batch4 & 59.5 & \textbf{63.9} & \textbf{55.3} & \textbf{59.6}    & \textbf{57.7}     &  &  \\
Batch5 & \textbf{59.8} & 62.7 & 54.1 & 58.9    & 57.1     &  &  \\ \cline{1-6}
\end{tabular}
\caption{\footnotesize Ablation results of randomly combining different batches of data. Batch1-Batch5 represent combinations of 0.25, 0.5, 0.75, 1 and 2 times the data as described in 3.2, respectively.} 
\label{table3}
\vspace{-1.0em}
\end{table}

\begin{table}[]
\resizebox{0.5\textwidth}{!}{
\begin{tabular}{c|ccccc}
\hline
\multirow{2}{*}{Method} & \multicolumn{5}{c}{Segment-level}                                             \\ \cline{2-6}
      & A             & V             & AV            & Type@AV       & Event@AV      \\ \hline
\multicolumn{1}{c|}{HAN}         & 60.1          & 52.9          & 48.9          & 54.0          & 55.4          \\
\multicolumn{1}{c|}{CMRC+HAN}    & \textbf{60.6} & \textbf{54.4} & \textbf{49.7} & \textbf{54.9} & \textbf{56.5} \\
\multicolumn{1}{c|}{MGN}         & 60.8          & 55.4          & 50.4          & 55.5          & 57.2          \\
\multicolumn{1}{c|}{CMRC+MGN}    & \textbf{61.3} & \textbf{57.5} & \textbf{53.0} & \textbf{57.3} & \textbf{58.4} \\
\multicolumn{1}{c|}{JoMoLD}      & 61.3          & 63.8          & 57.2          & 60.8          & 59.9          \\
\multicolumn{1}{c|}{CMRC+JoMoLD} & \textbf{62.3} & \textbf{64.8} & \textbf{57.9} & \textbf{61.7} & \textbf{61.0} \\ \hline
\multirow{2}{*}{Method} & \multicolumn{5}{c}{Segment-level}                                             \\ \cline{2-6}
     & A             & V             & AV            & Type@AV       & Event@AV      \\ \hline
\multicolumn{1}{c|}{HAN}         & 51.3          & 48.9          & 43.0          & 47.7          & 48.0          \\
\multicolumn{1}{c|}{CMRC+HAN}    & \textbf{51.8} & \textbf{50.0} & \textbf{43.0} & \textbf{48.2} & \textbf{48.9} \\
\multicolumn{1}{c|}{MGN}         & 51.1          & 52.4          & 44.4          & 49.3          & 49.1          \\
\multicolumn{1}{c|}{CMRC+MGN}    & \textbf{51.2} & \textbf{54.8} & \textbf{47.3} & \textbf{51.1} & \textbf{49.9} \\
\multicolumn{1}{c|}{JoMoLD}      & 53.9          & 59.9          & 49.6          & 54.5          & 52.5          \\
\multicolumn{1}{c|}{CMRC+JoMoLD} & \textbf{54.3} & \textbf{62.0} & \textbf{50.8} & \textbf{55.7} & \textbf{53.5} \\ \hline
\end{tabular}
}
\vspace{-0.5em}
\caption{\footnotesize Results of CMRC on different baselines.}
\label{table4}
\vspace{-2.0em}
\end{table}

\noindent \textbf{Mamba-based attention (MBA).} After extracting the features, we implement a simple segment-based attention mechanism by Mamba. This mechanism operates on the attention of ten frames or ten audio segments of each video, which effectively captures the salient features in a video and strengthens the feature representation of different segments. From the experimental results (\ref{table2}), it can be seen that all metrics show a decline when MBA is not used.


\noindent \textbf{Adaptive mamba fusion (AMF).} We introduce additional branch based on Vision Mamba \cite{thirty-three, thirty-eight} for dynamically adjusting the scanning order.  As shown in \ref{table2}, we can see that our approach achieves improvement. In the AVVP task, the events occurring in the two modalities are random, potentially being complementary or unrelated. Therefore,the AMF module shares cross-modal information while ensuring the independence of unimodal information. Besides, the addition of dynamic ordering branches mitigates the forgetting problem that mamba models pose in causal learning, which ignores information in early segments.


\noindent \textbf{Mamba feature enhancement module (MFE).} We design MFE to capture inter-modal similarities and enrich feature learning. The ablation experiment presented in \ref{table2} demonstrates the effectiveness of our proposed module. The metrics of the model are slightly improved after using MFE. The module amplifies the similar features of the two modalities at the same segment, thus allowing the two modalities to achieve a complementary effect in the prediction. We consider that when events occurring in both modalities are similar, one modality can assist in predicting events in the other modality (\eg hearing an engine and seeing a car). In the AVVP task, the similar features of two modalities have a greater probability of representing similar events, while the complementary features have a greater probability of representing different events. When the events occurring in the two modalities are not similar, noise tends to be introduced during modality interaction. MFE and AMF confirm the necessity to balance contribution in prediction by cross-modal interaction. 


\noindent \textbf{Pseudo-label semantic interaction module.} We introduce text modality to enhance the model's understanding of the scene. We encode segment-level pseudo-labels as text features that semantically interact with the corresponding visual/audio features. We use the text modality as a constraint to mitigate the noise interference caused by the other modality. As shown in \ref{table2}, it can be seen that all metrics of the model are improved after PLSIM. AMF, MFE and PLSIM exclude additional noise while retaining similar information across different modalities.

\noindent \textbf{Comparison with Transformer and CNN.} To more comprehensively demonstrate the capabilities of the proposed Mamba-Transformer model, we replace the Mamba component in AV-mamba with either Transformer (multi-head attention) or ResNet (Conv). In order to match AV-mamba, we use a single layer of multi-head attention or a single layer of ResNet in the visual and audio tracks respectively. As shown in \ref{table5}, MUG achieves better results.
\begin{table}[]
\resizebox{0.5\textwidth}{!}{
\begin{tabular}{c|cccccc}
\hline
\multirow{2}{*}{Method} & \multicolumn{5}{c|}{Segment-level}                                                            & \multirow{2}{*}{Parameters} \\ \cline{2-6}
                        & A             & V             & AV            & Type@AV       & \multicolumn{1}{c|}{Event@AV} &                             \\ \hline
CNN                    & 62.4              &  65.3             & 58.0              & 61.9              &  62.2                             &  \textbf{6.5M }                          \\
Transformer                     & 62.7          & 66.2          & 59.2          & 62.7          & 62.2                          & 19.3M                        \\
MUG                     & \textbf{65.4} & \textbf{66.5} & \textbf{59.9} & \textbf{63.9} & \textbf{64.7}                 & 7.6M                        \\ \hline
\multirow{2}{*}{Method} & \multicolumn{5}{c|}{Event-level}                                                              & \multirow{2}{*}{Parameters} \\ \cline{2-6}
                        & A             & V             & AV            & Type@AV       & \multicolumn{1}{c|}{Event@AV} &                             \\ \hline
CNN                     & 56.4              &  62.7             & 52.8              &  57.3             & 55.2                              &  \textbf{6.5M}                           \\
Transformer                     & 56.4          & 62.6          & 53.0          & 57.3          & 54.8                          & 19.3M                        \\
MUG                     & \textbf{59.5} & \textbf{63.9} & \textbf{55.3} & \textbf{59.6} & \textbf{57.7}                 & 7.6M                        \\ \hline
\end{tabular}
}
\vspace{-0.5em}
\caption{\footnotesize Comparison with Transformer and CNN.} 
\label{table5}
\vspace{-2.0em}
\end{table}


\section{Conclusion}
In this paper, we propose a pseudo labeling augmented audio-visual mamba network, which effectively enhances the model's capacity to learn from each segment. Data augmentation not only improves the quality of the dataset, but also allows model to acquire more fine-grained segment information. Additionally, a framework based on Mamba is proposed to enhance the perception ability both on single frame and multiple frames. Our approach proves its performance in a lot of experiments. Future work will focus on evaluating the effectiveness of MUG using larger datasets.